\documentclass[twoside,11pt]{article}

\usepackage{jmlr2e, xcolor}
\usepackage{microtype}
\usepackage{graphicx}
\usepackage{subfigure}
\usepackage{booktabs} 
\usepackage{amsmath}
\usepackage{amssymb}
\let\proof\relax 

\usepackage{amsthm}
\usepackage{algorithm}
\usepackage{algorithmic}
\usepackage{hyperref}
\usepackage{csquotes}

\newcommand{\N}{\mathbf{N}}
\newcommand{\R}{\mathbf{R}}

\newcommand{\e}{\epsilon}

\newcommand{\bb}{\mathbf{b}}
\newcommand{\dd}{\mathbf{d}}
\newcommand{\I}{\mathcal{I}}

\newcommand{\W}{\mathcal{W}}
\newcommand{\B}{\mathcal{B}}
\newcommand{\G}{\mathcal{G}}
\newcommand{\RR}{\mathcal{R}}
\newcommand{\s}{\mathcal{S}}
\newcommand{\wh}{\widehat}

\newtheorem{myremark}{Remark}[section]
\newtheorem{mydef}{Definition}[section] 
\newtheorem{mytheorem}{Theorem}[section]

\newtheorem{myconj}{Conjecture}[section]

\begin{document}

 \ShortHeadings{Multiparameter Persistence}{Coskunuzer et al.}
\firstpageno{1}

\title{Smart Vectorizations for Single and Multiparameter Persistence}

\author{\name Baris Coskunuzer \email coskunuz@utdallas.edu \\
       \addr Mathematical Sciences, University of Texas at Dallas, USA
       \AND
       \name Cuneyt Gurcan Akcora \email cuneyt.akcora@umanitoba.ca \\
       \addr  
       Computer Science, University of Manitoba, Canada
\AND
 \name Zhiwei Zhen \email zhiwei.zhen@utdallas.edu \\
\addr  
      Mathematical Sciences, University of Texas at Dallas, USA
\AND
 \name Ignacio Segovia Dominguez 
\email ignacio.segoviadominguez@utdallas.edu \\
 \addr Mathematical Sciences, University of Texas at Dallas, USA
\AND  \name Murat Kantarcioglu \email muratk@utdallas.edu \\
\addr  
      Computer Science, University of Texas at Dallas, USA
\AND  \name Yulia R. Gel \email ygl@utdallas.edu\\
\addr  
      Mathematical Sciences, University of Texas at Dallas, USA
}

\editor{}

\maketitle

\begin{abstract}
The machinery of topological data analysis becomes increasingly popular in a broad range of machine learning tasks, ranging from anomaly detection and manifold learning to graph classification.
Persistent homology is one of the key approaches here, allowing us to systematically assess the evolution of various hidden patterns in the data as we vary a scale parameter. The extracted patterns, or homological features,  along with information on how long such features persist throughout the considered filtration of a scale parameter, 
convey a critical insight into salient data characteristics and data organization.

In this work, we introduce two new and easily interpretable topological summaries for single and multi-parameter persistence,
namely, saw functions and multi-persistence grid functions, respectively. 
Compared to the existing topological summaries which tend to assess the numbers of topological features and/or their lifespans at a given filtration step, our proposed saw and multi-persistence grid functions allow us to explicitly account for essential complementary information such as the numbers of births and deaths at each filtration step.  

These new topological summaries can be regarded as the complexity measures of the evolving subspaces determined by the filtration and are of particular utility for applications of persistent homology on graphs. We derive theoretical guarantees on the stability of the new saw and multi-persistence grid functions and illustrate their applicability for graph classification tasks.
\end{abstract}

\section{Introduction}
\label{submission}

Topological data analysis (TDA) has been recently applied to many different domains ranging from cryptocurrency transaction analysis to anomaly detection.
Persistent homology-based approaches emerged as one of the key TDA techniques that allow systematical assessment of the evolution of various hidden patterns in the data as a function of a scale parameter. The extracted patterns, or homological features,  along with information on how long such features persist throughout the considered filtration of a scale parameter, 
convey a critical insight into underlying data properties.
In the past, such homological features have been successfully used as an input to various machine learning tasks to enable efficient and accurate models that are robust to noise in the data sets. 

In this work, we provide new and easily interpretable topological summaries for single and multi-parameter persistence, namely, saw functions and multi-persistence grid functions (MPGFs).  
Compared to the existing topological summaries which tend to assess the numbers of topological features and/or their lifespans at a given filtration step, our proposed saw functions allow us to extract additional information that captures the numbers of births and deaths at each filtration step. On the other hand, MPGFs are one of the first topological summaries in the Multi-Persistence case which needs no slicing. In our empirical evaluations, we show that these summaries are useful in building efficient and accurate graph classification models.

One important contribution of our work is to leverage these interpretable features to build accurate graph classification models even with limited training data. This is important because existing work on graph convolutional networks (GCNs) for graph classification require large amounts of data and expensive training process without providing any insights into why certain classification result is provided. Compared to GCN-based approaches, the proposed single and multi-parameter persistence features provide insights into changes in the underlying sets or graphs as the scale parameter varies and enables the building of an efficient machine learning model using these extracted features. Especially for the application domain where the size of the data set is limited, building machine learning models such as random forests using the topological summaries for single and multi-parameter persistence result in prediction accuracy close to best GCN models while providing insights into why certain features are useful.

Contributions of our paper can be summarized as follows:
\begin{itemize}
    \item We introduce two new topological summaries for 
    single- and multi-persistence: {\it Saw functions} and {\it Multi-persistence Grid Functions} (MPGFs). The new summaries are interpretable as complexity measure of the space 
    and, contrary to the existing descriptors, contain such salient complementary information on
    the underlying data structure as the numbers of births and deaths of topological features at each filtration step. 
    
    \item We prove theoretical stability 
    of the new topological summaries. Our MPGF is one of the first successful and provably stable vectorizations of multi-parameter persistence without slicing. The theory of multi-parameter persistence suffers from many technical difficulties to define persistence diagrams in general settings. Our MPGFs bypass these problems by going directly to the subspaces defined, and deliver 2D descriptors.
    \item The proposed new topological summaries are computationally simple and tractable.  
    Our MPGF takes a few minutes for most datasets, but on average provides results that differ as little as 3.53\% from popular graph neural network solutions.
    \item To the best of our knowledge, this is the first paper bringing the machinery of multi-persistence to graph learning. We discuss the utility and limitations of multi-persistence for graph studies.
    \end{itemize}

\section{Related Work}

{\bf Topological Summaries:} While constructing the right filtration for a given space $X$ is quite important, the vectorization and topological summaries of the information produced in persistent homology are crucial to obtain the desired results. This is because the summary should be suitable for ML tools to be used in the problem so that the produced topological information can address the question effectively. However, the produced persistent diagrams (PDs) are a multi-set that does not live in a Hilbert space and, hence, cannot be directly inputted into an ML model. 
In that respect, there are several approaches to solve this issue where they can be considered in two categories. The first one is called vectorization~\citep{di2015comparing,bubenik2012statistical, adams2017persistence} which embeds PDs into $\mathbb{R}^d$, and the second one is called kernelization~\citep{kusano2016persistence, le2018persistence, zielinski2019persistence, zhao2019learning} which is to induce kernel Hilbert spaces by using PDs. Both approaches enjoy certain stability guarantees and 
some of their key parameters are learnable.
However, the resulting performance of such topological summaries as classifiers is often highly sensitive to the choice of the fine-tuned kernel and transformation parameters.
In this paper, we find a middle way where our topological summaries both keep most of the information produced in PDs and do not need fine calibration.

In \citep{chung2019persistence}, the authors bring together the class of vectorizations which are single variable functions in the domain of thresholds in a nice way and call them Persistence curves. Here, the infrastructure is very general, and most of the current vectorization can be interpreted under this umbrella. In particular, Betti functions, life span, persistence landspaces, Betti entropy, and several other summaries can be interpreted in this class. 
Another similar infrastructure to describe the topological summaries of persistent homology especially in graph case is PersLay~\citep{carriere2020perslay}. In this work, the authors define a general framework for vectorizations of persistent diagrams, which can be used as a neural network layer. The vectorizations in this class are very general and suitable to include kernels in construction. 

{\bf TDA for graph classification:}
Recently, TDA methods have been successfully applied to the graph classification tasks often in combination with DL and learnable kernelization
of PDs~\citep{hofer2017deep, togninalli2019wasserstein,rieck2019persistent,le2018persistence, zhao2019learning, hofer2019learning, kyriakis2021learning}. Furthermore, as mentioned above, \citet{carriere2020perslay} unified several current approaches to PD representations to obtain the most suitable topological summary with a learnable set of parameters, under an umbrella infrastructure. 
Finally, most recently, \cite{cai2020understanding} obtained successful results by combining filtrations with different vectorization methods.

\section{Background} 
\label{sec-background}

In this section, we provide a brief introduction to persistent homology. Homology $H_k(X)$ is an essential invariant in algebraic topology, which captures the information of the $k$-dimensional holes (connected components, loops, cavities) in the topological space $X$. Persistent homology is a way to use this important invariant to keep track of the changes in a controlled topological space sequence induced by the original space $X$. For basic background on persistence homology, see \citep{edelsbrunner2010computational,zomorodian2005computing}. 

\subsection{Persistent Homology} \label{sec-PH1}

For a given metric space $X$, consider a continuous function  $f:X\to\R$. Define $X_{\alpha}=\{x\in X \mid f(x)\leq \alpha\}$ as the $\alpha$-sublevel set of $X$. 
Choose $\{\alpha_i\}$ as an increasing  sequence of numbers with $\alpha_0=\min f$ to $\alpha_N=\max f$. Then, $\{X_{\alpha_i}\}$ gives an exhaustion of the space $X$, i.e. $X_{\alpha_0}\subset X_{\alpha_1}\subset ...\subset X_{\alpha_N}=X$. By using $X_{\alpha_i}$ itself, or inducing natural topological spaces $\wh{X}_{\alpha_i}$ (e.g. VR-complexes), one obtains a sequence of topological spaces $\wh{X}_{\alpha_0}\subset \wh{X}_{\alpha_1}\subset \ldots \wh{X}_{\alpha_i}$ which is called a {\em filtration} induced by $f$. For each $i$, one can compute the $k^{th}$ homology group $H_k(\wh{X}_{\alpha_i})$ which describes the $k^{th}$ dimensional holes in $\wh{X}_{\alpha_i}$. The rank of the homology group $H_k(\wh{X}_{\alpha_i})$ is called the \textit{Betti number} $\B_k(\alpha_i)$ which is basically number of $k$-dimensional holes in the space $\wh{X}_{\alpha_i}$.

By using persistent homology, we keep track of the topological changes in the sequence $\{\wh{X}_{\alpha_i}\}$ as follows. When a topological feature $\sigma$ (connected component, loop, cavities) appears in $H_k(\wh{X}_{\alpha_i})$, we mark $b_\sigma=\alpha_i$ as its birth time. 
The rank of $H_k(\wh{X}_{\alpha_i})$ is called
{\em Betti number}. The feature $\sigma$ can disappear at a later time $H_k(\wh{X}_{\alpha_j})$ by merging with another feature or by being filled in. Then, we mark $d_\sigma=\alpha_j$ as its death time. We say $\sigma$ persists along $[b_\sigma,d_\sigma)$. The longer the length ($d_\sigma- b_\sigma$) of the interval, the more persistent the feature $\sigma$ is \citep{adams2021geometric}.  
A multi-set $PD_k(X,f)=\{(b_\sigma,d_\sigma)\mid \sigma \in H_k(\wh{X}_{\alpha_i})\}$ is called {\em persistence diagram} of $(X,f)$ which is the collection of $2$-tuples marking birth and death times of $k$-dimensional holes $\{\sigma\}$ in $\wh{X}_{\alpha_i}$. Similarly, we call the collection of the intervals $\{[b_\sigma,d_\sigma)\}$ {\em the barcodes} of  $(X,f)$, which store the same information with the persistence diagrams.

Even though the constructions are very similar, depending on the initial space $X$, and the filter function $f$, the interpretation and the meaning of the topological changes recorded by the persistent homology for $(X,f)$ might be different. For example, when the initial space $X$ is a point cloud, and $f(p)=d(p,X)$ is the distance function in the ambient space, this sublevel construction coarsely captures the shape of the point cloud as follows: The sublevel sets $f(p)=d(p,X)\leq \e$ gives a union of $\e$-balls $\bigcup_{x\in X}N_\e(x)$, and it naturally induces relevant complexes (Vietoris-Rips, Cech, Clique) to capture the coarse geometry of the set. In this case, by the celebrated Nerve Theorem, we observe that persistence homology and our filtration captures the shape of the underlying manifold structure of the point cloud \citep{edelsbrunner2010computational}. 

On the other hand, when we use a graph $G$ as our initial space $X$, and $f$ as a suitable filter function on $G$ (e.g. degree), the sublevel filtration gives a sequence of subgraphs (or complexes induced by them). In particular,  the filter function $f$ positions the nodes in a suitable direction and determines the direction for the evolution of these subgraphs. In this sense, the filter function is highly dominant to govern the behavior of the filtration, and outcome persistent homology. Hence, one can express that these outputs not only describe the topological space $X$'s features (or shape) but also the behavior of the filter function $f$ on $X$. While in the point cloud case, the persistent homology can be said that it captures the shape of the point cloud coarsely, in the graph case, the meaning of the outputs of the persistent homology is highly different. In the following section, we discuss the interpretation of the persistent homology induced by these different filter functions as measuring the topological complexity of these subspaces.

While the persistence homology originated with a single filter function, one can generalize the filtration idea to several functions. In particular, single parameter filtration can be viewed as chopping off the manifold into slices and bringing these slices together in an ordered fashion. For topologists, this is the Morse Theoretic point of view to analyze a given space $X$ by using a filter function (Morse function) $f$ \citep{mischaikow2013morse}. This approach can naturally be adapted to several filter functions. By slicing up space with more than one filter function at the same time, we can get a much finer look at the topological changes in the space in different filtering directions. This is called \textit{Multiparameter Persistence}. By construction, multiparameter persistence gives more refined information on the topological space, and the theory has been growing rapidly in the past few years \citep{carriere2020multiparameter,vipond2020multiparameter,harrington2019stratifying}.

As a simple example for the construction of multiparameter persistence, consider two filter functions $f,g:X\to \R$ on the space $X$. Let $\I=(\alpha_i,\beta_j)$ be the partially ordered index set where $(\alpha_i,\beta_j)\leq(\alpha_{i'},\beta_{j'})$ if $\alpha_i\leq\alpha_{i'}$ and $\beta_{j}\leq\beta_{j'}$. Then, define $X_{ij}=\{x\in X\mid f(x)\leq \alpha_i \mbox{ and } g(x)\leq \beta_j\}$. Similarly, for each fixed $i_o$, we get a single parameter filtration $\wh{X}_{i_o1}\subset\wh{X}_{i_o2} \subset ... \subset \wh{X}_{i_om_2}$ where $\wh{X}_{ij}$ is the complex induced by $X_{ij}$. Multiparameter persistence enables one to look at the space from more than one direction, and to detect topological changes with respect to more than one filter function at the same time. The advantage of multiparameter persistence over doing single parameter persistence multiple times is that it gives a finer filtration to understand the space, and one can further detect the interrelation between the filtering functions.

\subsection{Interpretation of Persistent Homology as Complexity Measure} \label{sec-PH2}

In the previous section, we discussed that the interpretation of persistent homology output highly depends on the filter function $f:X\to \R$. When $f(p)=d(p,X)$ is the distance function to space $X$, persistent homology captures the coarse shape of the point cloud given in ambient space. This point cloud mostly comes from a dataset embedded in a feature space. Then persistent homology gives information on patterns, topological features of the manifold structure where the dataset lies. Therefore, this output gives crucial information on the data and enables one to see the hidden patterns in it.

Even though in the origins of the persistent homology, the filtration constructed was induced by the distance function described above, the technique to construct persistent homology allows using a different type of filter functions.
When we choose a different filter function on the space, we get a completely different filtration, and naturally, the persistent homology output completely changes. Over the years, the technique of persistent homology highly developed, and generalized in many different settings. Depending on the question at hand, one applies suitable filter functions to space and gets a filtration to find the features and patterns detected by the topological tools.

For example, when the filter function $f:X\to\R$ is coming from a qualitative property of the set $X$, we can interpret the persistent homology output as the records of \textquote{complexity changes} in the space with respect to this filter function. This is because, in this setting, the persistent homology keeps track of the topological changes in the sequence of subspaces $\{\wh{X}_\alpha\}$ determined by the filter function. As $\alpha$ evolves, the topological complexity of $X_\alpha$ changes. One can describe topological complexity as the number of components, loops, cavities at a given threshold level $\alpha$. In this setting, birth and death times of topological features are not determined by the position of the points in $X$ with respect to each other, but they are determined by the filter function. This new persistent homology output does not solely depend on the space $X$, but mostly depends on the filter function $f$. Hence, one can also interpret the persistent homology as valuable information on the behavior of the filter function $f$ on $X$.

For instance, when $X$ is a graph $G$, and $f$ is the degree function on the vertices of $G$, then the persistent homology first describes the topological changes in the subgraphs generated by low degree nodes, and then evolves toward subgraphs with high degree places. The persistent homology output is the information on how the complexity changes in the filtration while the degree function increases.

With the motivation of this interpretation of persistent homology, we present two novel topological summaries capturing the \textquote{complexity changes} notion described above very well as easily interpretable single variable and multi-variable functions.

\section{Novel Topological Summaries: Saw Functions}	

In the following, we interpret persistent homology as a tool to keep track of the topological complexity changes in $X$ with respect to filter $f:X\to\R$, and propose two new topological summaries. We define our first topological summary, {\em the Saw Function}, for the single parameter case.

\subsection{Saw Functions} 

Saw Function counts the existing live topological features in the intervals $(\alpha_i,\alpha_{i+1}]$ similar to the Betti functions \citep{edelsbrunner2010computational}. The advantage of Saw Functions is that they remedy the loss of several essential information when vectorizing persistent diagrams. 

When passing from persistent diagrams to Betti functions, one loses most of the information on birth and death times as it only counts the number of live features at time $\alpha_i$. The death/birth information for the whole interval $(\alpha_i,\alpha_{i+1})$ is lost during this transition. With Saw Functions, we capture this information by keeping the number of deaths and births in the interval $(\alpha_i,\alpha_{i+1})$ in a natural way.

Throughout the construction, we use sublevel filtration as above, but the same construction can be easily adapted to superlevel, or similar filtrations. Given $(X,f)$ and $\I=\{\alpha_i\}$, let $\{\wh{X}_{\alpha_i}\}$ be the sublevel filtration as described in Section~\ref{sec-PH1}, i.e. $\wh{X}_{\alpha_0}\subset \wh{X}_{\alpha_1}\subset..\subset \wh{X}_{\alpha_N}$. 	

\begin{figure}
    \centering
    \includegraphics[width=0.6\linewidth]{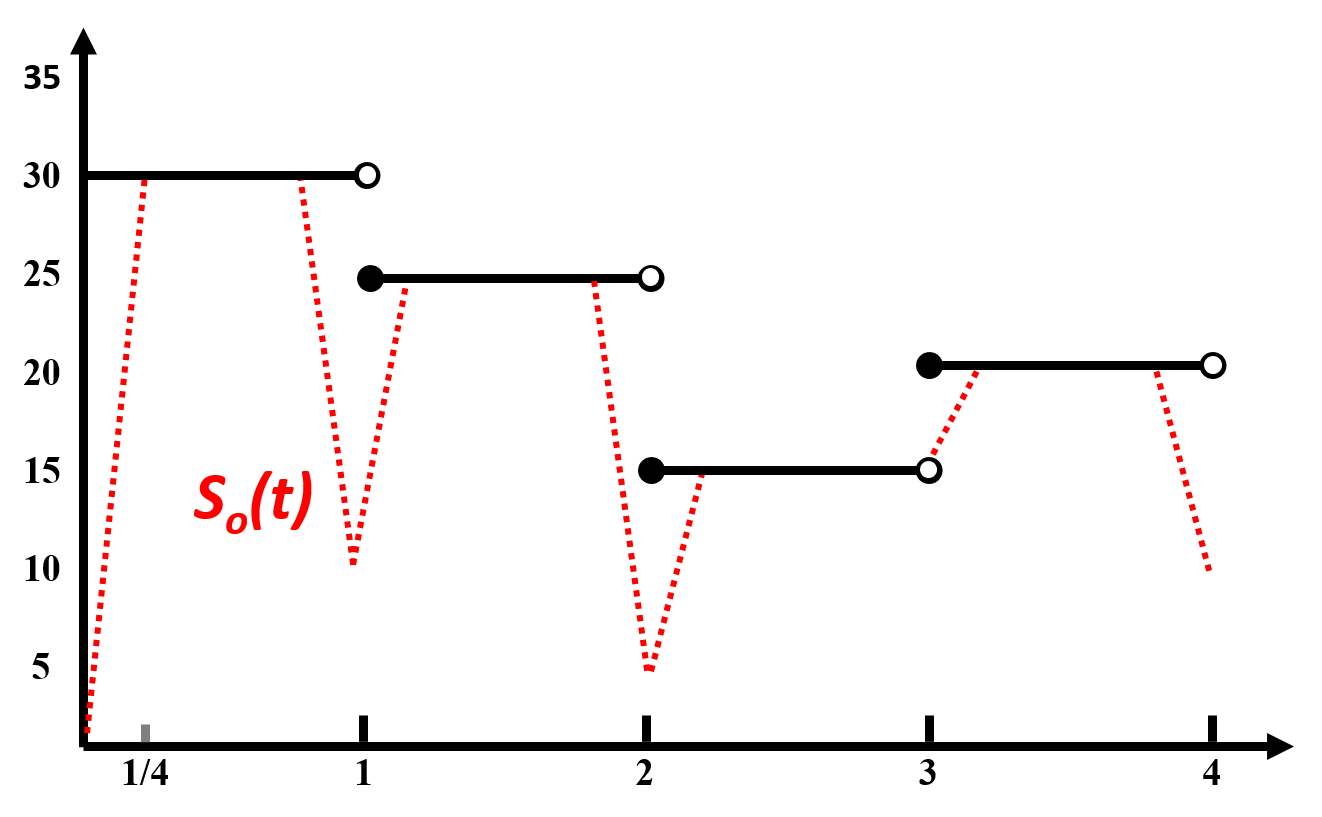}
    \caption{ $\B_0(t)$ is a step function with $\B_0(1)=30$, $\B_0(2)=25$, $\B_0(3)=15$, and $\B_0(4)=20$. We get $\s_0(t)$ by modifying $\B_0(t)$ with birth and death times $\bb(n),\dd(n)$. Here, $\bb(0)=30$, $\bb(1)=15$,  $\bb(2)=10$, $\bb(3)=5$, $\dd(1)=20$, $\dd(2)=20$,  $\dd(3)=0$ and $\dd(4)=10$.}
   \label{fig:Betti}
\end{figure}

Now, let $\B_k(\alpha_i)$ be the $k^{th}$ Betti number of $\wh{X}_{\alpha_i}$. This defines a function $\B_k:[\alpha_0,\infty)\to \N$ called $k^{th}$ \textit{Betti function} of the filtration $(X,f, \I)$ as follows: $\B_k(t)=\B_k(\alpha_j)$ where $\alpha_j$ is the largest number in $\I$ with $\alpha_j\leq t$. By definition, $\B_k$ is a piecewise constant function which only changes values at $\I$. See black function in Figure~\ref{fig:Betti}.

Let $PD_k(X)$ be the persistence diagram of $(X,f)$ at $k^{th}$ level. Here, we give the construction of Saw Function $\s_0(t)$ for $k=0$. Higher levels will be similar.

Let $PD_0(X)=\{(b_j,d_j)\}$ where $b_j,d_j$ represent the birth and death of $j^{th}$ component in the filtration. Notice that $b_j,d_j\in \I=\{\alpha_i\}$ for any $j$. Let $\bb(\alpha_i)=\sharp \{b_j=\alpha_i\}$ and $\dd(\alpha_i)=\sharp \{d_j=\alpha_i\}$, i.e. $\bb(\alpha_i)$ is the number of births  at $t\in(\alpha_{i-1},\alpha_i]$, and   $\dd(\alpha_i)$ is the number of deaths at $t\in(\alpha_{i-1},\alpha_i]$. The regular Betti functions only capture the information $\bb(\alpha_i)-\dd(\alpha_i)=\B_k(\alpha_i)-\B_k(\alpha_{i-1})$. With Saw Functions, we keep the number of births and deaths in the intervals $(\alpha_{i-1},\alpha_i]$, and embed it naturally in the function as follows.

When defining Saw Functions, we define a simple generator function for each point $(b_j,d_j)\in PD_0(X)$. Let $\chi_j=\chi(I_j)$ be the characteristic function for half closed interval $I_j=[b_j,d_j)$, i.e. $\chi_j(t)=1$ for $t\in I_j$, and $\chi_j(t)=0$ otherwise. Notice that $\B_0(t)=\sum_j\chi_j$ by construction.
	
To include the counting of birth and death times information in the function, we modify $\chi_j$. In general, for a given filtration $(X,f, \I)$, we choose a suitable {\em lag parameter} as follows:

Assuming the thresholds $\{\alpha_i\}$ are somewhat evenly distributed as in generic case, we let
$\lambda \sim \frac{1}{4}\mbox{average}(\{|\alpha_{i+1}-\alpha_i|\})$.  The reason for the coefficient $1/4$, and how to choose $\lambda$ in general case is explained in Remark \ref{rmk:lambda}. For simplicity, we assume our thresholds are integer valued with $\alpha_{i+1}-\alpha_i=1$. One can think of $X$ is a graph $G$, and $f$ is the degree, or eccentricity functions on the vertices of $G$. For this case, set $\lambda=1/4$. Then, modify $\chi_j$ as follows.

	$$\wh{\chi}_j(t)=\left\{  
	\begin{array}{cc}
	4(t-b_j) & t \in [b_j,b_j+\frac{1}{4}] \\
	4(t-d_j) & t \in [d_j-\frac{1}{4},d_j] \\
	\chi_j(t) & \mbox{otherwise}

	\end{array} \right.
	$$

	Now, define the Saw Function by adding up each generator $\wh{\chi}_j$ for each $(b_j,d_j)\in PD_0(X)$ as follows: $\s_0(t)=\sum_j \wh{\chi}_j(t)$. For general $k$, we have $$\s_k(t)=\sum_{(b^k_j,d^k_j)\in PD_k(X)} \wh{\chi}^k_j(t),$$ where $PD_k(X)$ is the $k^{th}$ persistent diagram of $X$.

	Notice that with this definition, we have a piecewise linear function $\s_k(t)$ with the following properties. For $n\in \I\subset \N$, 
	
	\begin{itemize}
	\item $\s_k(t)=\B_k(t)$ for $t\in [n+\frac{1}{4},n+\frac{3}{4}]$
	
	\item	$\B_k(n)-\s_k(n)=\bb_k(n)$ where $\bb_k(n)$ the number of births of $k$-cycles at time $t \in (n-1,n]$.
	
	\item $\B_k(n-1)-\s_k(n)=\dd_k(n)$ where $\dd_k(n)$ the number of deaths of $k$-cycles at time $t \in (n-1,n]$.
	\end{itemize}
	
Now, when the filtration function is integer valued with $\lambda={1}/{4}$, the explicit description of Saw Function $\s_k(t)$ is as follows. For any $n\in\I$ and $t\geq \alpha_0$, we have 

	$$\s_k(t)=\left\{ 
\begin{array}{ll}
\B_k(t)-4\dd_k(n)(t\mbox{-}n)+1 & t \in [n-\frac{1}{4},n) \\
\B_k(t)-\bb_k(n)+4\bb_k(n)(t\mbox{-}n) & t \in [n,n+\frac{1}{4}] \\
\B_k(t) & \mbox{ otherwise} \\
\end{array} \right.
$$

The novelty of Saw Functions is that it transforms most of the information produced by persistence diagrams in a simple way to interpretable functions. As described earlier, if one considers persistent homology as machinery which keeps track of the complexity changes with respect to the filter function, Saw Functions immediately summarizes the output in an interpretable way where one immediately reads the live topological features at the given times as well as the number of births and deaths at a given instance.

\begin{myremark} [Choice of the Lag Parameter $\lambda$] \label{rmk:lambda} 
In our sample construction, we chose $\lambda \sim \frac{1}{4}\mbox{average}(\{|\alpha_{i+1}-\alpha_i|\})$ with the assumption of the thresholds $\{\alpha_i\}$ are evenly distributed. The reason for the coefficient $1/4$ should be clear from the Figure \ref{fig:Betti}. Lag parameter $\lambda>0$ marks two points in the interval $(\alpha_i, \alpha_{i+1})$, e.g. $\alpha_i<\alpha_i+\lambda<\alpha_{i+1}-\lambda<\alpha_{i+1}$. Notice that on the interval $[\alpha_i+\lambda,\alpha_{i+1}-\lambda]$, our Saw functions coincides with the Betti functions. We chose $\lambda=1/4$ in our setting to make it the length ($\alpha_{i+1}-\alpha_i-2\lambda$) of the interval $[\alpha_i+\lambda,\alpha_{i+1}-\lambda]$ is the same with the length ($2\lambda$) of the interval where Betti function and the Saw function differs $(\alpha_i-\lambda,\alpha_i+\lambda)$, i.e. $\alpha_i+\lambda,\alpha_{i+1}-\lambda=2\lambda$. Depending on the thresholds set $\{\alpha_i\}$, and the problem at hand, one can choose varying lag parameters $\{\lambda_i\}$, too, i.e. $(\alpha_i-\lambda_i,\alpha_i+\lambda_i)$.
\end{myremark}

\subsection{Comparison with Similar Summaries:}

In the applications, since we also embed birth and death information into the function, when comparing persistent homologies of different spaces, Saw Functions as their vectorization will give us a finer understanding of their differences, and produce finer descriptors. In particular, if we compare the Saw Functions with Betti functions or Persistence Curves, and similar summaries \citep{umeda2017time,chen2020measuring} the main differences are two-fold: In non-technical terms, the summaries above are piecewise-constant and do not store the information on the birth/death times in the threshold intervals $(\alpha_i,\alpha_{i+1})$. However, embedding this information to Saw Functions give more thorough information around the topological changes near the threshold instances. If the number of births and deaths are high numbers at a threshold instance, Betti functions and similar summaries see only their differences. However, in Saw Functions, such high numbers around a threshold instance can be interpreted as an anomaly (or high activity) in the slice $f^{-1}((\alpha_i,\alpha_{i+1}))$ and the deep zigzag pieces exactly quantifies this anomaly. 

In technical terms, we can describe this difference as follows: Let $(X^+,f^+,\I^+)$ and $(X^-,f^-,\I^-)$ describe two filtrations. For simplicity, assume $\I^+=\I^-=\I$. Assume that they both induce the same Betti functions $\B_k^+(t)=\B_k^-(t)$ for $t\in [\alpha_0,\infty)$ in the $k^{th}$-level. Notice that this implies that $\bb^+_k(\alpha_i)-\dd^+_k(\alpha_i)=\bb^-_k(\alpha_i)-\dd^-_k(\alpha_i)$ for any $i$ by definition. On the other hand, as Saw Functions store the number of birth and death information, we see that $\s^+_k(t)$ may be highly different from $\s^-_k(t)$ as follows.

To see the difference in $L^1$-metric,  let $d_1(f,g)=\int_{\alpha_0}^\infty |f(t)-g(t)|\,dt$ represent $L^1$-distance between $f$ and $g$. Then, we have $d_1(\s^+_k,\s^-_k)=\sum_{i=1}^N \lambda\cdot|\bb^+_k(\alpha_i)-\bb^-_k(\alpha_i)|=\sum_{i=1}^N \lambda\cdot|\dd^+_k(\alpha_i)-\dd^-_k(\alpha_i)|$. Here, the second equality comes from the fact that $\bb^+_k(\alpha_i)-\dd^+_k(\alpha_i)=\bb^-_k(\alpha_i)-\dd^-_k(\alpha_i)$ by assumption. In other words, while the same filtration induces exactly same Betti functions $\B^+_k(t)=\B^-_k(t)$, the induced Saw Functions can be very far away from each other in the space of functions, and this shows that they induce much finer descriptors. 

Similarly, from $L^2$-metric perspective, Saw Functions produce much finer information  than the similar summaries in the literature because of the slope information in the zigzags. In particular, Let $d_2(f,g)=\int_{\alpha_0}^\infty |f(t)-g(t)| + |f'(t)-g'(t)|\,dt$ be the $L^2$-distance. If we use the notation above, $(X^\pm,f^\pm,\I)$ induces exactly same Betti functions $\B^+_k(t)=\B^-_k(t)$, while $d_2(\s^+_k,\s^-_k)= \sum_{i=1}^N (\lambda+1)\cdot|\bb^+_k(\alpha_i)-\bb^-_k(\alpha_i)|$. Notice that the distance is much higher in $L^2$-metric, as the lag parameter $\lambda$ can be a small number. In practice, if number of births and deaths near a threshold, $\bb_k(\alpha_i)$ and $\dd_k(\alpha_i)$, are both very close large numbers, Betti functions and similar summaries does not see this phenomena as they only detect their difference, $\bb_k(\alpha_i)-\dd_k(\alpha_i)$. However, in Saw Functions, we observe a steep slope, and deep zigzag near the threshold $\alpha_i$ which can be interpreted as "high activity", or "anomaly" near the slice $f^{-1}((\alpha_{i-1},\alpha_i))$, which can be an effective tool for classification purposes. 

\noindent {\em Tension at a threshold:} Motivated by the idea above, we define \textit{$k$-tension} of $\alpha_i$ as $\tau_k(\alpha_i)=\bb_k(\alpha_i)+\dd_k(\alpha_i)$ which represents the amount of the activity (total number of births and deaths) in the interval $(\alpha_{i-1},\alpha_i]$. As discussed above, while betti functions only detect the difference $\bb_k(\alpha_i)-\dd_k(\alpha_i)$, Saw Functions detect also their sum $\tau_k(\alpha_i)=\bb_k(\alpha_i)+\dd_k(\alpha_i)$ at a given threshold. In particular, the tension $\tau_k(\alpha_i)$ is the stepness of the zigzag at $\alpha_i$ as it is the sum of the slopes of the Saw Function at that point.

Here, we only discuss the comparison of Saw Functions with Betti functions, but similar arguments can also be used to compare other persistence curves or similar vectorizations. 

Another advantage of Saw Functions over other common summaries is that one does not need to choose suitable parameters or kernels in the process. In other summaries, one needs to make suitable fine assumptions on the parameters, or kernel functions which can highly distort the output if not done properly.

\noindent {\bf Saw Functions as Persistence Curves:} 

If we consider Saw Functions in the Persistence Curves Framework \citep{chung2019persistence}, 
our Saw Functions can be interpreted as a new Persistence Curve with the following inputs. With their notation, our Saw Function $\s_k(t)$ corresponds to the persistence curve with generating function $\psi(b,d,t)=\wh{\chi}_{(b,d)}(t)$ (defined in the proof above) and summary statistics $T=\sum$ (summation), i.e. $\s_k(t)=\sum_{(b,d)\in PD}\wh{\chi}_{(b,d)}(t)$. Note that with this notation the Betti function $\B_k(t)=\sum_{(b,d)\in PD}\chi_{(b,d)}(t)$ where $\chi_{(b,d)}(t)$ is the usual characteristic function for interval $[b,d)$, i.e. $\chi_{(b,d)}(t)=1$ when $t\in [b,d)$ and $\chi_{(b,d)}(t)=0$ otherwise.

\subsection{Stability of Saw Functions} 

In this part, we discuss the stability of Saw Functions as vectorizations of persistence diagrams. In other words, we compare the change in persistence diagrams with the change they cause in Saw Functions.

Let $(X^+,f^+,\I^+)$ and $(X^-,f^-,\I^-)$ define two filtrations. Let $PD_k(X^\pm)$ be the corresponding persistence diagrams in $k^{th}$ level. Define $p^{th}$ Wasserstein distance $\W_p$ between persistence diagrams as follows. We use $\W_p(X^+,X^-)$ notation instead of $\W_p(PD_k(X^+),PD_k(X^-))$ for short. Let $PD_k(X^+)=\{q_j^+\}\cup \Delta^+$ and  $PD_k(X^-)=\{q_l^-\}\cup \Delta^-$ where 
$\Delta^\pm$ represents the diagonal with infinite multiplicity. Here, $q_j^+=(b^+_j,d_j^+)\in PD_k(X^+)$ represents the birth and death times of a $k$-dimensional cycle $\sigma_j$.
Let $\phi:PD_k(X^+)\to PD_k(X^-)$ represent a bijection (matching). With the existence of the diagonal $\Delta^\pm$ in both sides, we make sure the existence of these bijections even if the cardinalities $|\{q_j^+\}|$ and $|\{q_l^-\}|$ are different. Then, $$\W_p(X^+,X^-)= \min_{\phi}(\sum_j\|q_j^+-\phi(q_j^+)\|_\infty^p)^\frac{1}{p}, \quad p\in \mathbb{Z}^+.$$

In turn, the bottleneck distance is $\W_\infty(X^+,X^-)=\max_j \|q_j^+-\phi(q_j^+)\|_\infty$.

We compare the Wasserstein distance between persistence diagrams with the distance between the corresponding Saw Functions. Let $\s_k^\pm(t)$ be the $k^{th}$ Saw Functions as defined above. Assuming $\s^\pm=0$ outside their domains $I^\pm$, the $L^p$-distance between these two functions defined as $$d_p(\s^+,\s^-)=\biggl(\int_{I^+\cup I^-}|\s^+(t)-\s^-(t)|^p\ dt\biggr)^\frac{1}{p}.$$

Now, we state our stability result:

\begin{mytheorem} \label{thm:sawstability} Let $(X^\pm,f^\pm,\I^\pm)$ be as defined in previous paragraph. Then, Saw Functions are stable for $p=1$, and unstable for $p=\infty$. i.e.
$$d_1(\s^+(t),\s^-(t)) \leq C\cdot \W_1(X^+,X^-).$$

\end{mytheorem}


\noindent {\em Proof:} 
For simplicity, we first prove the result for Betti functions, and modify it for Saw Functions. Notice that $\B_k(t)=\sum_i \chi_i$ and $\s_k(t)=\sum_i \wh{\chi}_i$, where the index $i$ represents the points in $PD_k(G)=\{q_i\}$ where $q_i=(b_i,d_i)$. That is, $\chi_i$ is the characteristic function for the interval $[b_i,d_i)$. Then, 
\begin{equation} \label{eqn1}
|\B_k^+(t)-\B_k^-(t)|\leq |\sum_i\chi^+_i-\sum_j\chi_j^-|\leq \sum_i|\chi^+_i-\chi^-_{\phi(i)}|
\end{equation}

Notice that  given $q^+=(b^+,d^+)$ and $q^-=(b^-,d^-)$, assuming $d^-\leq d^+$, there are two cases for $\chi^+-\chi^-$. In case 1, $d^-\leq b^+$, which gives $\int_{\alpha_0}^\infty |\chi^+-\chi^-|=|d^+-b^+|+|d^--b^-|\leq |d^+-d^-|+|b^+-b^-|$. In case 2, $d^-> b^+$ and we get $\int_{\alpha_0}^\infty |\chi^+-\chi^-|=|d^+-d^-|+|b^+-b^-|$. Hence, 
		\begin{eqnarray} \label{eqn2} 
\int_{\alpha_0}^\infty |\chi^+_i(t) &-& \chi^-_{\phi(i)}(t)|dt  \leq  |b_i^+-b_{\phi(i)}^-|+|d_i^+-d_{\phi(i)}^-|\ \nonumber \\  &\leq & 
	2\max\bigl\{|b_i^+-b_{\phi(i)}^-|,|d_i^+-d_{\phi(i)}^-|\bigr\} \nonumber \\ &= &2\|q^+_i-q^-_{\phi(i)}\|_\infty.
\end{eqnarray}
Now, in view of~(\ref{eqn1}), we obtain 
\begin{eqnarray} \label{eqn3}
d_p(\B_k^+(t),\B_k^-(t))&=&\biggl(\int_{\alpha_0}^\infty |\B_k^+(t)-\B_k^-(t)|^p \ dt \biggr)^\frac{1}{p}  \\ &\leq&
	 \biggl(\int_{\alpha_0}^\infty \bigl[\ \sum_i|\chi^+_i(t)-\chi^-_{\phi(i)}(t)|\ \bigr]^p \ dt \biggr)^\frac{1}{p} \nonumber
\end{eqnarray}
		
By~(\ref{eqn2}), the stability result for $p=1$ follows from
\begin{eqnarray*}
d_1(\B_k^+(t),\B_k^-(t))&=&\int_{\alpha_0}^\infty \sum_i|\chi^+_i(t)-\chi^-_{\phi(i)}(t)| dt\\ &\leq & \sum_i 2\|q^+_i-q^-_{\phi(i)}\|_\infty \\&=&2\W_1(X^+,X^-).
\end{eqnarray*}
		
This finishes the proof for Betti functions. Notice that Equation \ref{eqn2} is still true when we replace $\chi^\pm_i$ with $\wh{\chi}_i^\pm$. Then, when we replace 
Betti functions $\B^\pm_k(t)$ with $\s^\pm_k(t)$ and $\chi^\pm_i$ with $\wh{\chi}_i^\pm$ in the inequalities above, we get the same result for Saw Functions, i.e. $d_1(\s_k^+(t),\s_k^-(t))\leq 2\W_1(X^+,X^-)$. The proof follows. \hfill $\Box$

Note that when $p>1$, Equation \ref{eqn3} above may no longer be true.  So, the stability for $1<p<\infty$ is not known. For $p=\infty$, we have the following counterexample.

\noindent {\bf Counterexample  for $p=\infty$ Case:} 

Here, we describe a counterexample showing that similar statement for $p=\infty$ cannot be true, i.e. $d_\infty(\s^+(t),\s^-(t)) \leq C\cdot \W_\infty(X^+,X^-)$.
We define two persistence diagrams, where one has $n$ off-diagonal elements, and the other has no off-diagonal element. Let $PD_k(X^+)=\{(1,2)^n\}\cup\Delta$, i.e. $(1,2)$ has multiplicity $n$. Let $PD_k(X^-)=\Delta$, i.e. only diagonal elements. Note that it is straightforward to get such a $X^\pm$ for any $k$ and $n$. Then, as $d_\infty ((1,2),\Delta)=\sqrt{2}/2$, we have $\W_\infty(X^+,X^-)=\sqrt{2}/2$. On the other hand, each copy of the point $(1,2)$ contributes $1$ to betti function in the interval $[1,2)$. This means $\B^+_k(t)=n$ for $t\in [1,2)$ while $\B^-_k(t)=0$ for any $t$.  This gives $d_\infty(\s_k^+,\s_k^-)=\sup_t|\s_k^+(t)-\s_k^-(t)| =n$ while $\W_\infty(X^+,X^-)=\sqrt{2}/2$. This shows that there is no $C>0$ with $d_\infty(\s_k^+(t),\s_k^-(t))\leq C\cdot \W_\infty(X^+,X^-)$ in general.

\section{Multi-Persistence Grid Functions (MPGF)}

In this part, we introduce our second topological summary: \textit{Multi-Persistence Grid Functions} (MPGF). Like in Saw Functions, we aim to use persistent homology as a complexity measure of evolving subspaces determined by filtration. In this case, instead of one filter function, we use multiple filter functions. One can consider this approach as a vectorization of multiparameter persistent homology.

First, to keep the setup simple, we construct MPGFs in $2$-parameter setting. Then, we give the natural generalization in any number of parameters. Here, we only consider sublevel filtrations for more than one function at the same time. 

\subsection{MPGFs for 2-parameter}  

\begin{figure}
    \centering
    \includegraphics[width=0.6\linewidth]{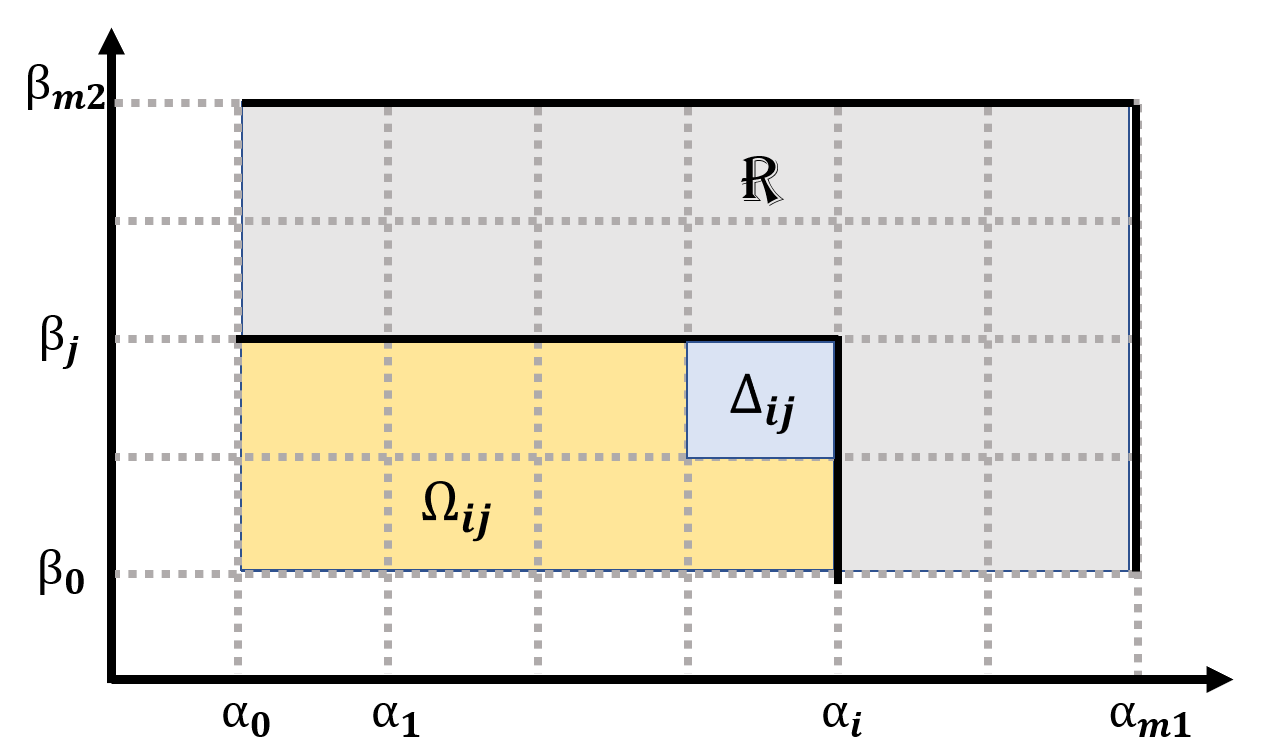}
    \caption{\textbf{Filtration.} The rectangle $R$ is the domain of $\G_k$ which consists of $m_1\times m_2$ small rectangles $\{\Delta_{ij}\}$. $\G_k$ is constant for each $\Delta_{ij}$ which is the $k$th Betti number of the complex coming from $F^{-1}(\Omega_{ij})$.}
    \label{fig-multi} 
\end{figure}

{\em A Short Outline:} Before starting the formal construction, we start by giving an outline of the setup. Given two filter functions $f,g:X\to \R$ with their threshold sets $\{\alpha_i\}$ and $\{\beta_j\}$ respectively, one can chop off $X$ into finer pieces  $X^{ij}=\{x\in X\mid f(x)\leq \alpha_i, \ g(x)\leq \beta_j\}$. The subspaces $\{X^{ij}\}$  induce a natural filtration, however partially ordered nature of the threshold set $\{(\alpha_i,\beta_j)\}$, getting persistence homology in the multiparameter setting is highly technical \citep{lesnicklecturenotes2019}. We bypass this technicalities, by directly going to the subspaces $\{X^{ij}\}$, and considering their Betti numbers $\B_k(X^{ij})$ as the subspaces' topological complexity measure. Then, by using the grid $\{(\alpha_i,\beta_j)\}$ in the threshold domain $R$, we assign these Betti numbers to the corresponding boxes $\Delta_{ij}$ as in Figure \ref{fig-multi}. This construction defines a $2$-parameter function $\G_k(x,y)$ which we call {\em MPGF}.

Next is the formal construction of MPGFs.
Let $X$ be a metric space, and  $f,g: X\to\R$ be two filter functions. Define $F:X\to \R^2$ as $F(x)=(f(x), g(x))$. Let $\I_1=\{ \alpha_1,...\alpha_{m_1}\}$ be the set of thresholds for the first filtration $f$, where $\alpha_1=\min f(x)<\alpha_2 ...<\alpha_{m_1}=\max f(x)$. 
Similarly, $\I_2=\{\beta_1,...,\beta_{m_2}\}$ is defined for $g$. Let $\delta_1={(\alpha_{m_1}-\alpha_1)}/{(m_1-1)}$, $\delta_2={(\beta_{m_2}-\beta_1)}/{(m_2-1)}$, $\alpha_{0}=\alpha_{1}-\delta_1$, and $\beta_{0}=\beta_{1}-\delta_2$.
We define our MPGFs on the rectangle $\RR=(\alpha_0, \alpha_{m_1}]\times (\beta_0,\beta_{m_2}]$. Notice that the points $\{(\alpha_i,\beta_j)\}$ gives a grid in our rectangle $R$. 

Let $\Delta_{ij}=(\alpha_{i-1},\alpha_{i}]\times (\beta_{j-1},\beta_j]$ for $1\leq i\leq m_1$ and $1\leq j\leq m_2$. Then, $\RR=\bigcup_{i,j}\Delta_{ij}$.
Let $\Omega_{ij}=(\alpha_0,\alpha_i]\times (\beta_0,\beta_j]$ (see Figure \ref{fig-multi}).

Define  $X^{ij}=\{x\in X\mid f(x)\leq \alpha_i \mbox{ and } g(x)\leq \beta_j\}$, i.e. $X^{ij}=F^{-1}(\Omega_{ij})$. Then, let $\wh{X}^{ij}$ be the complex generated by $X^{ij}$. $\wh{X}^{ij}$ can be the set $X^{ij}$ itself, or VR, or similar complex induced by $X^{ij}$. Then, for each fixed $i_0$, $\wh{X}^{i_01}\subset \wh{X}^{i_02}\subset...\subset \wh{X}^{i_0m_2}$ gives a single parameter filtration for $(X^{i_0},g)$ where $X^{i_0}=f^{-1}((-\infty,\alpha_{i_0}])$.
 
\begin{mydef} \label{defn:MPGF} Let $\B_k(\wh{X}^{ij})$ be the $k^{th}$-Betti number of $\wh{X}^{ij}$. Define $k^{th}$ MPGF of the filtration for the filter function $F:X\to \R^2$ as
$$\G_k(x)=\B_k(\wh{X}^{ij})$$ for any $x\in \Delta_{ij}$.
\end{mydef}

That is, we assign every rectangle $\Delta_{ij}$ of the threshold domain, we assign the Betti number of the space $\wh{X}^{ij}$ corresponding to these thresholds. Notice that $\G_k:\RR\to \N$ has constant value at any small rectangle $\Delta_{ij}\subset \RR$. The algorithm is outlined in Alg.~\ref{alg:multi}.

\noindent \textbf{Complexity:} Algorithm~\ref{alg:multi} requires computation of Betti functions at each of the $m_1\times m_2$ grid cell. Betti computations require finding the rank of the homology group, which has a complexity of $\tilde{O}(|A|+r^\omega)$ where $|A|$ is the number of non-zero entries in the group, r is the true rank of the group and $\omega < 2.38$ is the multiplication
exponent \citep{cheung2013fast}.

\begin{figure}[h]
	\begin{center}
		\centerline{\includegraphics[width=0.7\linewidth]{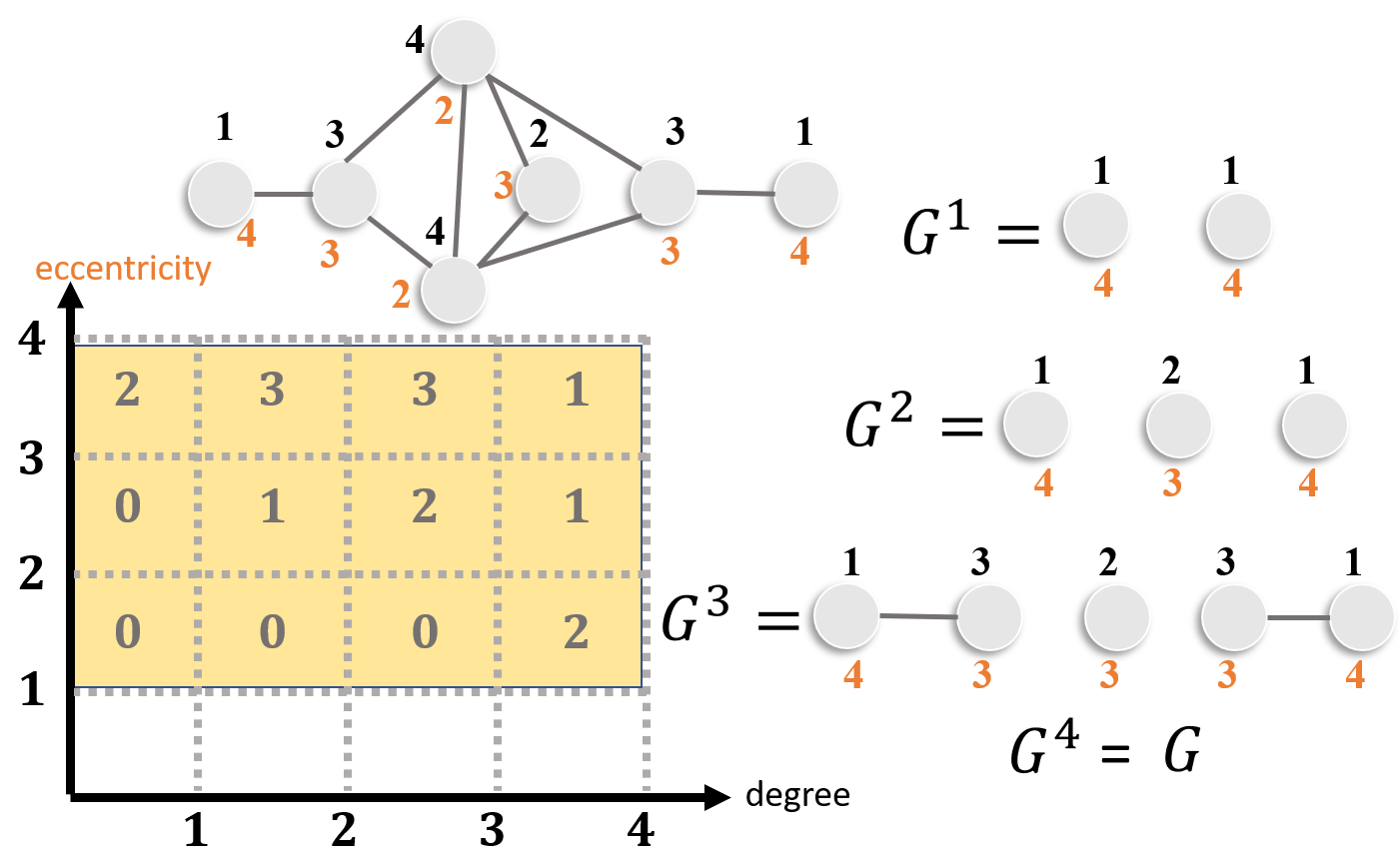}}
		\caption{\footnotesize \textbf{Grid.} In the graph $G$, black numbers are node degrees whereas red numbers are  node eccentricities (i.e., max distance to any node in the graph). In the right, $G^i$ represents the subgraph corresponding to sublevel set $f(v)\leq i$ where $f$ is the degree function. The subgraph $G^{ij}$ of $G^i$ is defined by $g(v)\leq j$ where $g$ is the eccentricity function.  MPGF $\G_0(\Delta_{ij})=\B_0(G^{ij})$.}
		\label{fig-multi_example}
	\end{center}
\end{figure}

\noindent {\bf Example:} In Figure \ref{fig-multi_example}, we described a simple example to show how MPGF works. In the example, $X$ is graph $G$, and $f,g:V\to\N$ integer valued functions on vertices. Here, $\wh{X}^{ij}$ is the subgraph $G^{ij}$ generated by $V^{ij}=\{v\in V\mid f(v)\leq i, \ g(v)\leq j\}$. In the grid, we count the number of topological feature exist in $X^{ij}$, and record this number to the box $\Delta_{ij}=(\alpha_{i-1},\alpha_{i}]\times (\beta_{j-1},\beta_j]$. In the example, we see MPGF $\G_0$ for $0$-cycles (connected components). 

\noindent {\bf Why MPGFs are useful?} 
Multiparameter filtration slices up space $X$ in a much finer and controlled way than single parameter filtration so that we can see the topological changes with respect to both functions at the same time.  Then, by using MPGFs for the corresponding multiparameter persistence, we can analyze, and interpret both functions' behavior on the space $X$, and interrelation between these two functions. In many cases, we have more than one important descriptors that are not directly connected, but both give very valuable information on the dataset. In that case, when both functions are used for multiparameter persistence, then together both functions will provide much finer information on the dataset, and the functions' undisclosed relationship with each other could be observed. One very popular case is the time-series datasets. One function is the time, and the other is another important qualitative descriptor of the dataset.

\begin{algorithm}[ht]
   \caption{Multi-Persistence Graph Classification}
   \label{alg:multi}
\begin{algorithmic}
   \STATE {\bfseries Input:} graph $G$, functions $f, g$, grid sizes $m_1,m_2$.
   \STATE {\bfseries Output:} A matrix of betti 0 and 1 values defined for a grid filtration.
   \STATE Compute $f$ values for $\forall v \in G$
   \STATE Fix $\alpha_i$ for $1\leq i\leq m_1$.
   \STATE Initialize matrix $grid$
   \FOR{$1\leq i\leq m_1$}
   \STATE Define $G^i=\{v\in \G\mid f(v)\leq \alpha_{i}\}$.
    \STATE Set $G^i$ as the original input space for the filtration function $g: X^i\to \R$
    \FOR{$1\leq j\leq m_2$}
   \STATE Set subgraph  $G^{ij} = \{v\in G^i \mid f(v)\leq \beta_j\}$.
   \STATE $grid[i,j,0]= \B_0(G^{ij})$
   \STATE $grid[i,j,1]= \B_1(G^{ij})$
   \ENDFOR
   \ENDFOR
   \STATE RETURN $grid$
\end{algorithmic}
\end{algorithm}

\subsection{MPGFs for any number of parameters} 

We generalize the idea above in any number of parameters as follows: Let $F:X\to \R^d$ be our filter function. In particular, let $F=(f_1,f_2,..,f_d)$. Let $\I_i=\{\alpha_{ij}\mid 1\leq j\leq m_i\}$ be the thresholds of the filter function $f_i$. In particular, $\{\alpha_{i_0j}\}$ is the thresholds of $f_{i_0}$ where $\alpha_{i_01}=\min_V f_{i_0}(v)<\alpha_{i_{02}}<..<\alpha_{i_0m_{i_0}}=\max_V f_{i_0}(v)$. 

Again, let $\delta_i={(\alpha_{im_i}-\alpha_{i1})}/{(m_i-1)}$, and let $\alpha_{i0}=\alpha_{i1}-\delta_i$. Define $d$-dimensional rectangular box $$\RR=(\alpha_{10},\alpha_{1m_1}]\times(\alpha_{20},\alpha_{1m_2}]\times...\times(\alpha_{d0},\alpha_{dm_d}].$$

Notice that the $d$-tuples $\{(\alpha_{1j_1},\alpha_{2j_2},...\alpha_{dj_d})\}$ deliver a $d$-dimensional grid in $R$ where $1\leq j_i\leq m_i$. Similarly, define small $d$-dimensional boxes \begin{eqnarray*}
    \Delta_{j_1j_2..j_d}&=&(\alpha_{1(j_1-1)}, \alpha_{1j_1}] \\ &\times& (\alpha_{2(j_2-1)}, \alpha_{2j_2}]\times \ldots \times (\alpha_{d(j_d-1)}, \alpha_{dj_d}].
\end{eqnarray*}

Then, again we have $\RR=\bigcup_{j_1,..j_d}\Delta_{j_1j_2..j_d}$. In other words, $\RR$ is the union of $N=m_1\cdot m_2\cdot ...\cdot m_d$ small boxes.

Similarly, define the subspace $X^{j_1j_2..j_d}=\{x\in X\mid f_i(x)\leq \alpha_{ij_i} \mbox{ for } 1\leq i\leq d\}$ of $X$ induced by the thresholds $j_1,j_2,...,$ and $j_d$. Let $\wh{X}^{j_1j_2..j_d}$ be the complex induced by $X^{j_1j_2..j_d}$. Here, $\wh{X}^{j_1j_2..j_d}$ can be the subspace $X^{j_1j_2..j_d}$ itself, or the VR (or similar) complex induced by $X^{j_1j_2..j_d}$. 

Let $\B_k(\wh{X}^{j_1j_2..j_d})$ $k^{th}$ Betti number of the space $\wh{X}^{j_1j_2..j_d}$. Define $k^{th}$ MPGF of the filtration $F:X\to \R^d$ as
$$\G_k(x)=\B_k(\wh{X}^{j_1j_2..j_d})$$ for any $x\in \Delta_{j_1j_2..j_d}$. Notice that $\G_k:\RR\to \N$ has constant value at any small box $\Delta_{j_1j_2..j_d}\subset \RR$. 

\noindent {\bf Algorithm for MPGFs in any number of parameters: }
We fix $j_1,...,j_{d-1}$ for $1\leq j_i\leq m_i$. Define $X^{j_1j_2..j_{d-1}}=\{x\in X\mid f_i(v)\leq \alpha_{ij_i} \mbox{ for } 1\leq i\leq d-1\}$. Then, set $X^{j_1j_2..j_{d-1}}$ as the original input space for the filtration function $f_d: X^{j_1j_2..j_{d-1}}\to \R$. Compute the
persistence diagrams, and corresponding Betti functions for this filtration. Then, in particular $\wh{X}^{j_1j_2..j_{d-1}\wh{j}}$ is the $\wh{j}^{th}$ complex in this filtration.  $\G_k(\Delta_{j_1j_2..j_{d-1}\wh{j}})=\B_k(\wh{X}^{j_1j_2..j_{d-1}\wh{j}})$, $k^{th}$-Betti number of $\wh{X}^{j_1j_2..j_{d-1}\wh{j}}$. In other words, if one can obtain $\wh{X}^{j_1..j_d}$, and compute its $k^{th}$-Betti number directly, then $\G_k(\Delta_{j_1...j_d)}=\B_k(\wh{X}^{j_1...j_d})$. 

\subsection{Comparison with other Multiparameter Persistence Summaries} 

The crucial advantage of our MPGFs over the existing topological summaries of the multiparameter persistence is its simplicity and generality. Except for some special cases, Multiparameter Persistence theory suffers from the problem of the nonexistence of barcode decomposition because of the partially ordered structure of the index set $\{(\alpha_i,\beta_j)\}$ \citep{thomas2019invariants,lesnick19}. The existing approaches remedy this issue by slicing technique by studying one-dimensional fibers of the multiparameter domain. In particular, in the threshold domain $\Omega=[\alpha_0,\alpha_N]\times[\beta_0,\beta_M]$ in the $xy$-plane, one needs to find a good slice (line segment) $ax+by=c$ in $\Omega$, where the multipersistence restricted to such slice can be read as a single parameter persistence. This makes the summaries vulnerable to the choice of "good slicing" (choice of $a,b,c$) and combining back the multiple single parameter information into a multi-parameter setting. For example, in \citep{carriere2020multiparameter}, the authors define a similar grid object called "Multiparameter Persistence Image" where they slice the domain $\RR$, and consider restricted single parameter persistence diagrams in the slices. Then, they successfully remedy the nonexistence of the barcode decomposition problem by matching the nearby barcodes to the same family, which they call \textit{vineyard decomposition}. Then, by using this, for each grid point, they take a weighted sum of each barcode family to see the \textquote{topological importance} of that grid point in this decomposition. They use 3 types of fibering (vineyard family) for the domain $\Omega$, which are expected to give good results for a generic case. However, their construction heavily depends on these slicing choices, and when filtration functions do not interact well, different slicing might give different images.

In our MPGFs, we take a more direct approach to construct our summaries. We bypass the serious technical issues in defining multiparameter persistence diagrams and directly go to the topological behavior of the subspaces induced by each grid. This gives a simple and general method that can be applied to any topological space with more than one filtration function. The topological summary is much simpler, and interpretable as one can easily read the changes in the graph. It also enables one to read the interrelation between the filtration functions, and their behavior on the space. Furthermore, as it does not use any slicing, it directly captures the multidimensional information produced by filtration functions.

\noindent {\bf Life Span Information:} 
Notice that in both Saw Functions and Multi-Persistence Grid functions, when vectorizing persistence diagrams, we consider persistence homology as complexity measurement during the evolution of filter function. While these functions give the count of live topological features, they seem to miss important information from persistence diagrams, i.e. life spans. Life span $d_\sigma-b_\sigma$ of a topological feature $\sigma$ is crucial information to show how persistent the feature is. In particular, if $d_\sigma-b_\sigma$ is large, it is an essential/persistent feature, while if short, it can be considered nonessential/noise. In that sense, our topological summaries do not carry the individual life span information directly, but indirectly. In particular, any persistent feature $\sigma$ contributes to Saw or MPGF functions as many numbers of intervals (boxes) they survive.

\subsection{Stability of MPGFs} 
	
In this part, we discuss the stability of MPGFs. Even though the multiparameter persistence theory is developing very fast in the past few years, there are still technical issues to overcome to use this technique in general. As mentioned before, because of the partially ordered nature of the threshold domain in the multiparameter case, the single parameter persistent homology theory does not generalize immediately to the multiparameter case. Similarly, distance definitions have related problems in this setting. There are several distance functions (interleaving, matching) suggested in the literature for multiparameter persistence \citep{thomas2019invariants}.  The theory is very active, and there are several promising results in recent years \citep{lesnick2015theory, carriere2020multiparameter}.  Here, we conjecture that our MPGFs are stable with respect to matching distance, and we sketch proof of the conjecture in a special case.

Let $(X^+,F^+,\I^+)$ and $(X^-,F^-,\I^-)$ define two filtrations where $F^\pm=(f^\pm,g^\pm)$, and $\I^\pm=\{(\alpha_i^\pm,\beta_j^\pm)\}$. Let $\mathfrak{D}_M(X^+,X^-)$ represent the\textit{ matching distance }of multiparameter persistence modules induced by $(X^+,F^+,\I^+)$ and $(X^-,F^-,\I^-)$ \citep{thomas2019invariants}.

Next, by assuming $\G^\pm=0$ outside of their domains $\RR^\pm$, define the $L^1$-distance between induced MPGFs as $$\mathbf{d}(\G^+,\G^-)=\int\int_{\RR^+\cup\RR^-} |\G^+(x,y)-\G^-(x,y)|\, dxdy$$

Now, we have the following conjecture for the stability of MPGFs.

\begin{myconj} Let $(X^\pm,F^\pm,\I^\pm)$ be as defined above. Then, MPGFs are stable i.e.
$$\mathbf{d}(\G^+,\G^-) \leq C\cdot \mathfrak{D}_M(X^+,X^-).$$
\end{myconj}

Here, the main motivation for this conjecture is that it naturally holds in the technically simplest case. Again, the partially ordered nature of the thresholds in multiparameter setting prevents having good barcode decomposition for the generators like in the single parameter persistent homology \citep{lesnicklecturenotes2019}. However, there are simple cases where good barcode decomposition exists \citep{lesnicklecturenotes2019}. In such a $2$-parameter setting, a barcode $\mathcal{B}^\pm_i$ is not one rectangle, but unions of rectangles (with vertices at thresholds) where the union is connected. Let $\{\mathcal{B}^\pm_i\}$ be the set of barcodes in our multiparameter persistent homology for $(X^\pm, F^\pm, \I^\pm)$. Then, the induced MPGFs $\G^\pm$ can be defined just like single parameter persistent homology as follows. In particular, let $\chi_{\mathcal{B}^\pm_i}(x,y)$  be the characteristic function for the barcode $\mathcal{B}^\pm_i$, then we have $$\G^\pm(x,y)=\sum_i\chi_{\mathcal{B}^\pm_i}(x,y).$$
In that case, our proof for Theorem \ref{thm:sawstability} goes through similar way by replacing generator functions $\wh{\chi}^\pm_j(t)$ in the proof with the characteristic functions $\chi_{\mathcal{B}^\pm_i}(x,y)$ of barcodes of multiparameter persistence. Since having a good barcode is a very special case, we leave this as a conjecture for general setting.

\subsection{An alternative definition for MPGFs} 

\begin{figure}
    \centering
    \includegraphics[width=.6\linewidth]{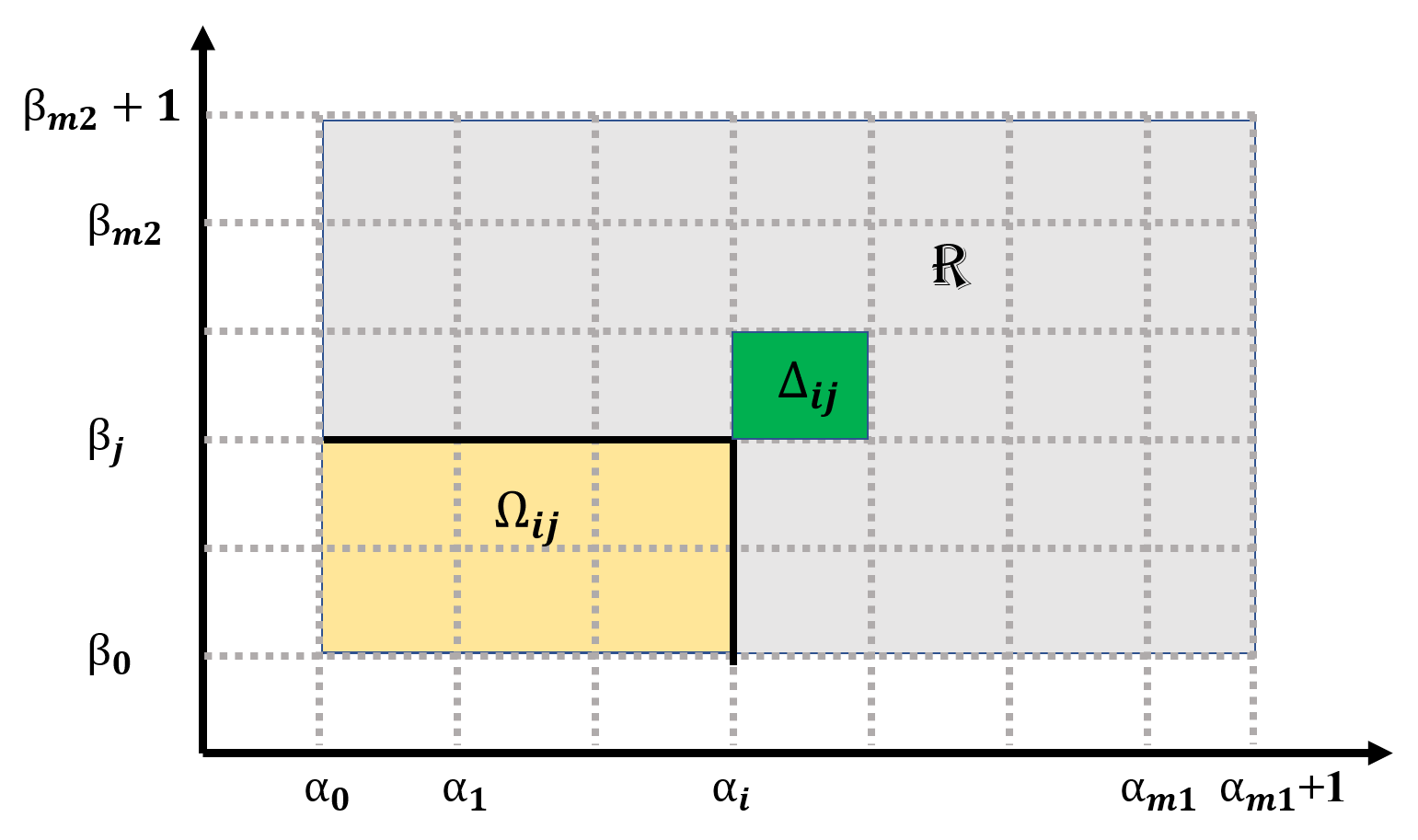}
    \caption{\footnotesize The rectangle $R$ is the domain of $\B_k$ consists of $m_1.m_2$ small rectangles $\{\Delta_{ij}\}$. $\B_k$ is constant for each $\Delta_{ij}$ which is the $k$ Betti number of the complex coming from $F^{-1}(\Omega_{ij})$.}
    \label{fig:fig-multi2}
\end{figure}

Even though MPGFs are giving the Betti numbers of corresponding subspaces $\{X^{ij}\}$, when restricted to single parameter, MPGFs are different than regular Betti functions. While single parameter Betti functions defined as $\B_k([\alpha_i,\alpha_{i+1}))=\B_k(\alpha_i)$, when MPGFs restricted to single parameter, we slide the interval one unit back so that $\G_k((\alpha_{i-1},\alpha_{i}])=\G_k(\alpha_i)$. Here, both $\G_k(\alpha_i)=\B_k(\alpha_i)$ represents the $k^{th}$ Betti number of the space $X^i=\{x\in X \mid f(x)\leq \alpha_i\}$.
We chose this presentation for MPGFs since it is more natural and easy to read for multiparameter case as it can be seen in Figure~2 in the main text.  
In particular, if one wants to define MPGFs consistent with the single parameter Betti functions, one needs to replace the half open intervals $(\alpha_i, \alpha_{i+1}]\times (\beta_j,\beta_{j+1}]$ in the original definition with the $[\alpha_i, \alpha_{i+1})\times [\beta_j,\beta_{j+1})$ starting from $\alpha_0=\min f$, and $\beta_0=\min g$, then the similar construction will give the definition for MPGFs consistent with the single parameter Betti functions. In this definition, the extra column goes to the right not to the left. Compare the Figure \ref{fig:fig-multi2} with the Figure~2 in the main text. This alternative definition has exactly the same information with our MPGFs, but as mentioned above, our definition is more user friendly in multiparameter case.

\noindent {\bf Betti Function Notation:} 
When restricted to single parameter, our MPGFs above is slightly different than regular Betti functions. While single parameter Betti functions defined as $\B_k([\alpha_i,\alpha_{i+1}))=\B_k(\alpha_i)$, in our notation, we slide the interval back so that $\B_k((\alpha_{i-1},\alpha_{i+1}])=\B_k(\alpha_i)$. This is because this presentation  is more natural and easy to read for multiparameter case as it can be seen in Figure \ref{fig-multi}. Recall that $\G_k(\Delta_{ij})=\B_k(X_{ij})$ where $X_{ij}=F^{-1}(\Omega_{ij})$. With our notation, we choose $\Delta_{ij}$ inside $\Omega_{ij}$ which is more natural in multiparameter case. However, if we want to define MPGF $\G_k$ as direct generalization of Betti function $\B_k$, then we need to use this alternative definition described above. Then, with this definition, $\Delta_{ij}$ will be outside of $\Omega_{ij}$ as in Figure \ref{fig:fig-multi2}.

\noindent {\bf Superlevel filtrations for MPGFs:} 

If the goal is to use superlevel filtration in original MPGFs, we
apply the following 
reverse approach: 

Let $X$ be a metric space, and  $f,g: X\to\R$ be two filtration functions. Let $F:X\to \R^2$ defined as $F(x)=(f(x), g(x))$. Let $\I_1=\{ \alpha_1,\ldots, \alpha_{m_1}\}$ be the set of thresholds for the first filtration $f$, where $\alpha_1=\min f(x)<\alpha_2 \ldots <\alpha_{m_1}=\max f(x)$. Let $\I_2=\{\beta_1,\ldots,\beta_{m_2}\}$ defined similarly for $g$. Now, let $\delta_1={(\alpha_{m_1}-\alpha_1)}/{(m_1-1)}$ and $\delta_2={(\beta_{m_2}-\beta_1)}/{(m_2-1)}$. Let $\alpha_{m_1+1}=\alpha_{m_1}+\delta_1$, and $\beta_{m_2+1}=\beta_{m_2}+\delta_2$.

Let $\Delta_{ij}=[\alpha_{i},\alpha_{i+1})\times [\beta_{j},\beta_{j+1})$ for $1\leq i\leq m_1$ and $1\leq j\leq m_2$. Then, $\RR=\bigcup_{i,j}\Delta_{ij}$.
Let $\Omega_{ij}=[\alpha_i,\alpha_{m+1})\times [\beta_j,\beta_{m_2+1})$. That is, in the sublevel filtration, we take lower left part of $\RR$ as $\Omega_{ij}$, while in superlevel filtration, we consider the upper right corner of $\RR$ as $\Omega_{ij}$. 

If  $X^{ij}=\{x\in X\mid f(v)\geq \alpha_i \mbox{ and } g(v)\geq \beta_j\}$, i.e. $X^{ij}=F^{-1}(\Omega_{ij})$, then $\wh{X}^{ij}$ is defined as the complex generated by $X^{ij}$.
Now, let $\B_k(\wh{X}^{ij})$ be the $k^{th}$-Betti number of $\wh{X}^{ij}$. The $k^{th}$ MPGF of the filtration $F:X\to \R^2$
is then given by
$\G_k(x)=\B_k(\wh{X}^{ij})$ for any $x\in \Delta_{ij}$.

A recursive algorithm to compute $\G_k$ can be described as follows. One can proceed by slicing $\RR$ inductively. Let $X^d_{j_d}=\{x\in X\mid f_d(x)\leq \alpha_{dj_d}\}$. Let $\wh{X}^d_{j_d}$ be the complex induced by $X^d_{j_d}$. This defines a $(d-1)$-dimensional slice in $\R$. Then, one can do $(d-1)$-parameter filtration on $\wh{X}^d_{j_d}$ with the filtration functions $\{f_1,..,f_{d-1}\}$. We now start the process with $X^d_{j_d}$ as original input space for $d-1$-parameter filtration, finish  the process inductively, and complete the function $\B_k$. Both methods computationally demonstrate same efficiency, but coding inductively in the second case could be easier.

\section{Application to Graph Classification}
In the following, as a case study, we will apply our topological summaries to the graph classification problem. Our task is to classify graph instances of a dataset. In particular, let $G$ be a graph with vertex set $V=\{v_i\}$ and edge set $E=\{e_{ij}\}$, i.e. $e_{ij}\in E$ if there is an edge between the vertex $v_i$ and $v_j$ in $G$. Let $f:V\to \R$ be a function defined on the vertices of $G$. $\I=\{\alpha_i\}$ be the threshold set which is an increasing  sequence with $\alpha_1=\min f$ to $\alpha_N=\max f$. Let $V^i=\{v\in V\mid f(v)\leq \alpha_i\}$. Let $G^i$ be the subgraph generated by $V^i$, i.e. $e_{ij}\in G^i$ if $v_i,v_j\in G^i$.  Then,  $G^1\subset G^2\subset...\subset G^N$ defines \textit{the sublevel filtration} for $(G,f)$. Similarly, one can define $\wh{G}^i$ as the clique complex of $G^i$. Then, $\wh{G}^1\subset \wh{G}^2\subset...\subset \wh{G}^N$ defines another natural filtration defined by $(G,f)$. We call this \textit{clique sublevel filtration}.

\subsection{Datasets}

In experiments, we used COX2~\citep{KKMMN2016}, DHFR~\citep{KKMMN2016}, BZR~\citep{KKMMN2016},
PROTEINS \citep{borgwardt2005protein}, NIC1~\citep{wale2008comparison} and DHFR~\citep{schomburg2004brenda} for binary and multi-class classification of chemical compounds, whereas IMDB-BINARY,
IMDB-MULTI, REDDIT-BINARY and REDDIT-5K~\citep{yanardag2015deep}
are social datasets. Table~\ref{tab:dataset} shows characteristics of the graphs.

\begin{table*}[h]
    \centering
    \begin{tabular}{l r r r r}
    \toprule
       Dataset & NumGraphs & NumClasses& AvgNumNodes & AvgNumEdges\\
       \midrule
       BZR & 405 & 2 &35.75 & 38.36\\
       COX2&467&2&41.22&43.45\\
       DHFR&467&2&42.43&44.54\\
FRANKENSTEIN& 4337&2&16.90&17.88\\
IMDB-BINARY & 1000 & 2 & 19.77 & 96.53\\
IMDB-MULTI & 1500 & 3 & 13.00 & 65.94\\
NCI1 & 4110 & 2 & 29.87 & 32.30\\
PROTEINS & 1113 & 2 & 39.06 & 72.82\\
REDDIT-BINARY & 2000 & 2 & 429.63 & 497.75\\
REDDIT-MULTI-5K & 4999 & 5 & 508.82 & 594.87\\
\bottomrule
    \end{tabular}
    \caption{Characteristics of the datasets used in experiments.}
    \label{tab:dataset}
\end{table*}

Reddit5K has five and ImdbMulti has three label classes. All other datasets have binary labels.

We adopt the following traditional node functions~\citep{hage1995eccentricity} in graph classification: 0) degree, 1) betweenness and 2) closeness centrality,  3) hub/authority score, 4) eccentricity, 5) Ollivier-Ricci and 6) Forman-Ricci curvatures. Ollivier and Forman functions are implemented in Python by Ni et al. \citep{ni2019community}. We compute functions 0-4 by using the iGraph library in R. 

We implement our methods on an Intel(R) Core i7 CPU @1.90GHz and 16Gb of RAM without parallelization. Our R code is available at \url{https://github.com/cakcora/PeaceCorps}.

\subsection{Models}
\label{sec:models}

In graph classification, we will compare our results to five well-known graph representation learning methods benchmarked in a recent in-depth study by Errica et al.~\citep{Errica2020A}. These are GIN~\citep{xu2018powerful},  DGCNN~\citep{zhang2018end}, DiffPool~\citep{ying2018hierarchical}, ECC~\citep{simonovsky2017dynamic} and GraphSage~\citep{hamilton2017inductive}.

For our results, we employ Random Forest~\citep{breiman2001random}, SVM~\citep{noble2006support}, XGBoost~\citep{chen2015xgboost} and KNN with Dynamic Time warping~\citep{muller2007dynamic} to classify graphs. Best models were chosen by 10-fold cross-validation on the training set, and all accuracy results are computed by using the best models on out-of-bag test sets.

Random Forest consistently gave the best accuracy results for both Single and Multi-Persistence experiments. SVM with Radial Kernel yielded the next best results. In the rest of this paper, we will demonstrate the Random Forest results. In Random Forest, we experimented with the number of predictors sampled for splitting at each node (mtry) from $\sqrt n-3,\ldots,\sqrt n+3$ where n is the number of dataset features. Our models use $ntree=500$ trees.

All results are averaged over 30 runs. In Single Persistence, we use Saw Signatures of length 100. In Multi-Persistence, 10x10 grids yielded the best results.    

We compute classification results on ten datasets, but only six of those have been used in the five benchmarked methods. As such, we report multi-persistence results for the five datasets in Table~\ref{tab:multiPersistence}, and rest of the datasets in ~\ref{tab:addGMultiPersistence}.

\subsection{Results}

\subsubsection{Single Persistence}
Single Persistence results are shown in Table~\ref{tab:singlePersistence}. Each row corresponds to a model that uses Saw Signatures from the associated Betti functions only. We report the best-performing filtration function for each dataset for Bettis 0 and 1. Betweenness and Closeness filtrations yield the most accurate classification results in four and five cases, respectively. Curvature methods  Ollivier and Forman are computationally costly, and generally yield considerably lower accuracy values. Although curvature methods yield the best results for Protein, the closeness filter reaches similar (i.e., $0.698$ and $0.695$) accuracy values for the dataset. We advise against using curvature methods due to their computational costs.

\begin{table}[h]
\caption{Single-persistence classification results for best  performing (in accuracy) filtrations. The \textit{ N(\%)} column of the table reports the percentage of graph instances that can be classified by using the associated model. Some Betti 1 models cannot classify all graph instances, because these graphs do not form Betti 1 values (i.e., 1-dimensional holes do not exist). }
\label{tab:singlePersistence}
\vskip 0.15in
\begin{center}
\begin{small}
\begin{sc}
\begin{tabular}{lcccc}
\toprule
dataset&filt&betti&acc&N(\%) \\
\midrule

IMDBBinary&betw&0&64.8 $\pm$ 3.5&100\\
IMDBBinary&betw&1&65.6 $\pm$ 2.7&100\\
IMDBMulti&clos&0&41.1 $\pm$ 3.4&100\\
IMDBMulti&clos&1&44.2 $\pm$ 3.8&98\\
NCI1&clos&0&69.8 $\pm$ 1.3&100\\
NCI1&clos&1&59.9 $\pm$ 1.8&82.7\\
Protein&ollivier&0&70.9 $\pm$ 3.2&100\\
Protein&forman&1&72.2 $\pm$ 3.1&100\\
Reddit5k&betw&0&47.8 $\pm$ 1.1&100\\
Reddit5k&clos&1&41.7 $\pm$ 1.8&91.8\\
RedditBinary&deg&0&86.0 $\pm$ 1.5&100\\
RedditBinary&betw&1&69.4 $\pm$ 2.0&86.0\\
\bottomrule
\end{tabular}
\end{sc}
\end{small}
\end{center}
\end{table}

\begin{table}[h]
\caption{Single-persistence classification results for best  performing filtrations (in accuracy) for additional datasets.}
\label{tab:leftoutsinglePersistence}
\vskip 0.15in
\begin{center}
\begin{small}
\begin{sc}
\begin{tabular}{lccrr}
\toprule
dataset&filt&betti&acc&N (\%) \\
\midrule

BZR&deg&0&84.3$\pm$3.5&100\\
BZR&deg&1&85.4$\pm$5.2&60.4\\
COX2&betw&0&77.4$\pm$3.3&100\\
COX2&ricci&1&82.3$\pm$4.6&76.1\\
DHFR&clos&0&79.8$\pm$2.0&100\\
DHFR&clos&1&70.7$\pm$3.4&95.6\\
FRANKENSTEIN&betw&0&67.0$\pm$1.3&100\\
FRANKENSTEIN&deg&1&72.1$\pm$7.2&4.4\\
\bottomrule
\end{tabular}
\end{sc}
\end{small}
\end{center}
\end{table}

\begin{figure}
    \centering
    \includegraphics[width=.7\linewidth]{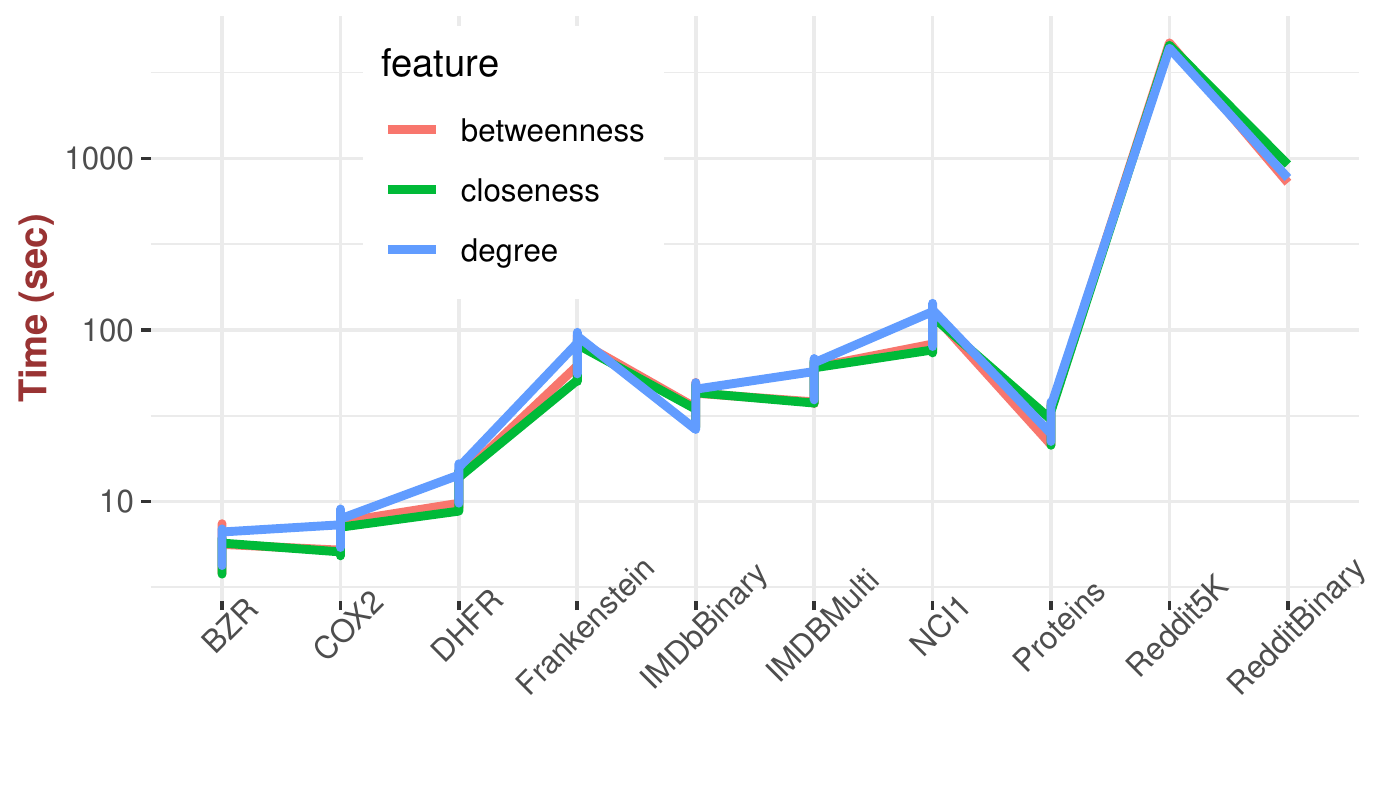}
    \caption{\textbf{Timings.} Single filtration costs of the three functions in experiments. Most dataset filtrations take a few minutes to complete. At most, 2000 RedditBinary and 4999 RedditMulti graphs take  4709 (betw) and 920 (clos) seconds, respectively. }
    \label{fig:timings}
\end{figure}

\begin{figure}[ht!]
    \centering
        \begin{subfigure}[]{\includegraphics[height=0.3\textheight,width=0.42\linewidth]{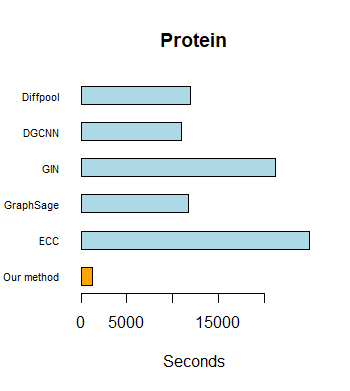}}\label{fig:Protein_single}\quad
         \end{subfigure}
         \begin{subfigure}[]{\includegraphics[height=0.3\textheight,width=0.42\linewidth]{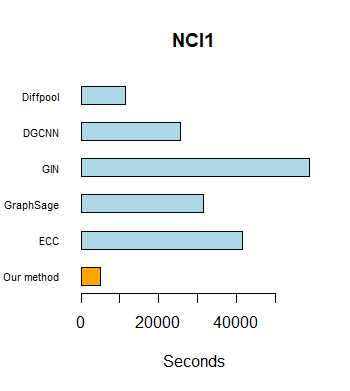}}\label{fig:NCI1_single}\quad
         \end{subfigure}
   \caption{Running time comparison results for two datasets. (a)Proteins (b) NCI1. For our method, the cost is for a complete end-to-end run. For other methods, the costs are for 10 epochs only. Note that in practice, these methods are run for 1000 epochs or more~\citep{Errica2020A}. Hence our end-to-end run times are considerably shorter than those of the five methods.}
   \label{running_time}
\end{figure}

\subsubsection{Multi Persistence}
 Multi-Persistence results are shown in Table~\ref{tab:multiPersistence} for best filtration grids. We achieved the best results over betweenness, closeness, and degree filtration pairs across all datasets.  On average our method provides accuracy that differs as little as 3.53\% from the five popular graph neural network solutions. Especially for the large Reddit graphs, our method ranks 3rd among the methods.

\begin{table}[h]
\caption{Multi-persistence classification accuracy results for best performing filtrations. }
\label{tab:multiPersistence}
\begin{center}
\begin{small}
\begin{sc}
\begin{tabular}{lccr|ccccc}
\toprule
\multicolumn{4}{c}{MPGFs}&DGCNN&GIN&DiffPool&ECC&GraphSage\\
ImdbB&deg&betw&67.8 $\pm$  2.7&70.0&71.23&68.4&67.67&68.80\\
ImdbM&betw&clos&44.3 $\pm$ 3.4&47.8&48.53&45.64&43.49&47.56\\
Nci1&deg&betw&74.0 $\pm$  1.6&74.4&80.04&76.93&76.18&76.02\\
Protein&betw&clos&73.8 $\pm$  2.8&75.5&73.25&73.73&72.30&73.01\\
Reddit5K&clos&betw&51.6 $\pm$  1.2&49.20&56.09&53.78&\textcolor{red}{OOR}&50.02\\
RedditB&deg&clos&89.0 $\pm$  0.9&87.7&89.93&89.08&\textcolor{red}{OOR}&84.32\\
\bottomrule
\end{tabular}
\end{sc}
\end{small}
\end{center}
\end{table}

\begin{table}[h]
    \centering
    \begin{tabular}{cccc}
    \toprule
    dataset&Filt1&Filt2&Acc\\
    \midrule
       BZR&deg&clos&84.3$\pm$3.5\\
    COX2&deg&betw&79.0$\pm$4.0\\
    DHFR&clos&betw&79.5$\pm$2.3\\
    FRANKENSTEIN&deg&betw&69.4$\pm$1.3\\
    \end{tabular}
    \caption{Additional results for multi-persistence classification for most accurate filtrations. These datasets are not studied by our competitors.}
    \label{tab:addGMultiPersistence}
\end{table}

A major advantage of our method is its low computational costs, compared to those of the five neural network solutions, which can take days. For example, as Table~\ref{tab:multiPersistence} shows, some methods run out of resources and cannot even complete the task. In comparison, as Figure~\ref{fig:timings} shows our method takes a few minutes to compute filtrations for all but two large Reddit networks. In Multi-Persistence, filtrations can be further parallelized for each row or column of the grid to reduce Multi-Persistence time costs.

\textbf{Comparison:} On average,  our MPGF takes a few minutes for most datasets, but on average provides results that differ as little as 3.53\% from popular graph neural network solutions with the caveat that Single Persistence with Betti 1 may not classify all graph instances. We demonstrate the additional results in Table~\ref{tab:leftoutsinglePersistence} for Single Persistence and Table~\ref{tab:addGMultiPersistence} for Multi Persistence for the BZR, COX2, DHFR and FRANKENSTEIN datasets. Except for the DHFR dataset, Betti 1 results are more accurate than Betti 0 results.

\section{Lessons Learned}

\noindent\textbf{Efficiency:} Single persistence yields lower (i.e., 1\%--6\%) accuracy values than Graph Neural Network solutions, but it is considerably faster to compute. In general, single persistence can be preferred when the classified graphs are too large to be handled with Graph Neural networks.

\noindent\textbf{Multi-Filtration:} In filtrations, we aimed to capture local (degree, eccentricity) and global (closeness and betweenness) graph structure around each node. We hypothesized that multi-persistence would benefit from looking at a graph from both local and global connections. This intuition proved correct, but we also observed that in some cases filtrations with global functions (e.g., betw and clos) reached the highest accuracy values (Table~\ref{tab:multiPersistence}). Especially in datasets with $\geq3$ classes, global connections may play a more important role.

\begin{figure}
    \centering
    \includegraphics[width=.7\linewidth]{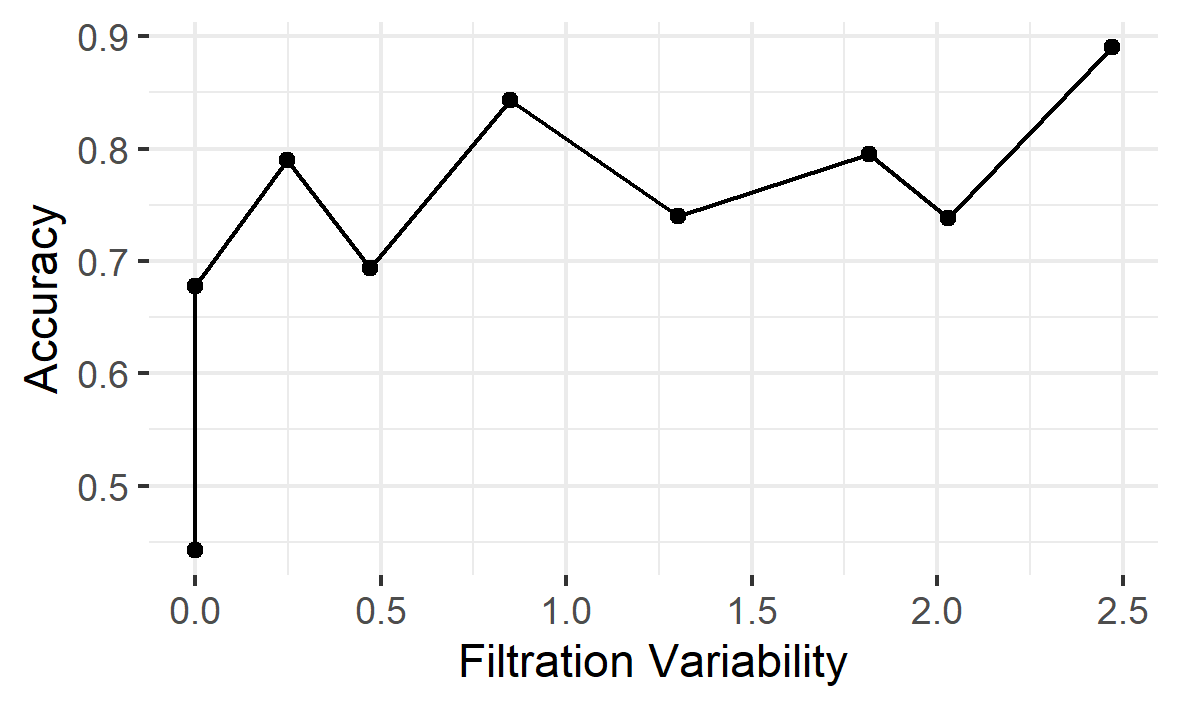}
    \caption{\textbf{Variability.} The impact of filter variability on accuracy. Low variability implies having the same view on the graph from both filters, which seems to affect performance negatively. The first two data points with 0 and $0.0007$ variability yield the lowest accuracy in MPGFs.}
    \label{fig:variance}
\end{figure}

\noindent\textbf{Variability is good:} In Multi-Persistence, filtrations may yield similar numbers of persistent features across thresholds. This is due to the dominant impact of edge connectivity in the graph, which causes similar persistent holes to form even when different filtrations are used. Consider the zero-dimensional holes created by two filters. We observe that when filters (e.g., betw, deg) create different numbers of zero-dimensional holes in all thresholds, MPFGs tend to learn models that reach higher accuracy. We compute a variability (variance) of Betti 0 numbers created by two filters to capture this behavior.   Figure~\ref{fig:variance} shows the relationship between variability and accuracy. As the variability increases, MPGFs can learn better models.
 
\section{Conclusion} 
In this paper, we have brought a new perspective to persistence homology as a complexity measure, and with this motivation, we have defined two new topological summaries in the single and multi-parameter setting. These summaries turn out to be very interpretable,  computationally efficient with a success rate comparable to the state-of-the-art models. We have only applied these summaries in a graph classification setting, but multi-persistence has enough promise to be useful in many complex problems.  

\bibliography{pacecorps.bib}
\end{document}